\documentclass{ao2e}%[seceqn,secfloat,secthm]

\usepackage{graphicx}
\usepackage[T1]{fontenc}
\usepackage{times}%
\usepackage{natbib}
\usepackage{mdframed}
\usepackage{color}
\firstpage{1}
\lastpage{3}
\volume{3}
\pubyear{2015}
\definecolor{orange}{rgb}{1,0.5,0}

\begin{document}
                          % The preamble begins here.
%
%\pretitle{Pretitle}
\title{FOCA: A Methodology for Ontology Evaluation}
\runningtitle{The FOCA Methodology}
%\subtitle{Subtitle}

\author[A]{\fnms{Judson} \snm{Bandeira}}%\thanks{Corresponding author: First Author}}
,
\author[A]{\fnms{Ig} \snm{Ibert Bittencourt}}
,
\author[B]{\fnms{Patricia} \snm{Espinheira}}
and
\author[C]{\fnms{Seiji} \snm{Isotani}}

\runningauthor{Bandeira, Bittencourt, Espinheira and Isotani}

\address[A]{UFAL, Av Lourival Melo Mota, S/N - Tabuleiro dos Martins, Macei\'o, Alagoas, Brazil\\
E-mail: \{jmb,ig.ibert\}@ic.ufal.br}

\address[B]{UFPE, Av. Professor Morais Rego, 1235 - Cidade Universit\'aria, Recife, Pernambuco, Brazil\\
E-mail: patespipa@de.ufpe.br}

\address[C]{USP, Av. Professor Almeida Prado, 1280, Butant\~a, S\~ao Paulo, S\~ao Paulo, Brazil\\
E-mail: sisotani@icmc.usp.br}

\begin{abstract}
Modeling an ontology is a hard and time-consuming task. Although methodologies are useful for ontologists to create good ontologies, they do not help with the task of evaluating the quality of the ontology to be reused. For these reasons, it is imperative to evaluate the quality of the ontology after constructing it or before reusing it. Few studies usually present only a set of criteria and questions, but no guidelines to evaluate the ontology. The effort to evaluate an ontology is very high as there is a huge dependence on the evaluator's expertise to understand the criteria and questions in depth.  Moreover, the evaluation is still very subjective. This study presents a novel methodology for ontology evaluation, taking into account three fundamental principles: i) it is based on the Goal, Question, Metric approach for empirical evaluation; ii) the goals of the methodologies are based on the roles of knowledge representations combined with specific evaluation criteria; iii) each ontology is evaluated according to the type of ontology. The methodology was empirically evaluated using different ontologists and ontologies of the same domain. The main contributions of this study are: i) defining a step-by-step approach to evaluate the quality of an ontology; ii) proposing an evaluation based on the roles of knowledge representations; iii) the explicit difference of the evaluation according to the type of the ontology iii) a questionnaire to evaluate the ontologies; iv) a statistical model that automatically calculates the quality of the ontologies.
\end{abstract}


\section{Introduction}
The word "ontology" has a wide range of definitions that can be found in the literature (\cite{r2}, \cite{r18}, \cite{r19} and \cite{r20}). Although many definitions have been proposed, Guarino (1998) provided a definition of the concept, which is widely accepted by the Applied Ontology community: 

\begin{center}
	\textit{An ontology is a logical theory accounting for the intended meaning of a formal vocabulary, i.e. its ontological commitment to a particular conceptualization of the world. The intended models of a logical language using such a vocabulary are constrained by its ontological commitment. An ontology indirectly reflects this commitment (and the underlying conceptualization) by approximating these intended models.}
\end{center}

It is very important to note that an ontology shares ontological commitment from other ontologies, otherwise it would be pointless to create an ontology.  This is why it is important, for those ontologies intended to support large-scale interoperability, to be well-founded, in the sense that the basic primitives they are built on are sufficiently well-chosen and axiomatized to be generally understood (defined by \cite{r21}). 

For this reason, ontologists need to fully understand the universe of discourse, knowing what the main concepts are in order to make the ontology more coherent with that universe of discourse. However, the higher the complexity of the universe of discourse is, the more the quality of the ontology tends to decrease.

For example, consider two ontologies, which model a bicycle and a car. To model the domain of the bicycle, the car tire concept was reused.  However, it is known that the size of the car tire is different (bigger) than the bicycle tire. Thus, the machine will understand that a car tire can also be used for a bicycle, which is not true. Considering this, the ontologist needs to include restrictions concerning the domain (in this case, the size of the tire) to avoid the problem of false agreement (the same tire is used by a bicycle and by a car). Figure 1 shows the problem of false agreement, where M(L) is "the world of vehicles", the dark ellipses are the correct models of the bicycle and car (with restrictions). If the ontologist does not impose the size restriction, the intended models of the bicycle (dark ellipse) will overlap the models of the car ontology (grey ellipse).

\begin{figure}[!htb]
	\centering
	\includegraphics{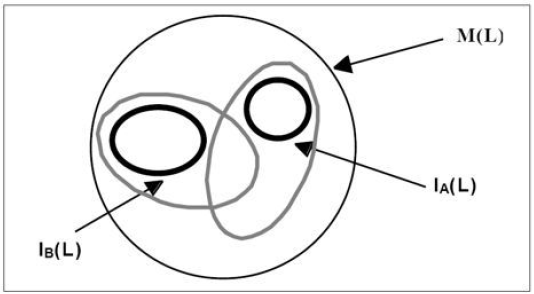}
	\caption{The problem of false agreement}
	\label{}
\end{figure}

As a result, modeling an ontology is a hard and time-consuming task. As a consequence of this difficulty, the worse the quality of the ontology is, the lesser is reusability. The best way to deal with such a difficulty is to use methodologies to build ontologies\footnote{ There are several methodologies available on the literature to build ontologies (\cite{r22} and \cite{r23}) }(just as philosophers do with categorization methods). Although methodologies are useful for ontologists to create good ontologies, they do not help with the task of deciding which ontology to reuse. Another very important point is that there are thousands of ontologies available on the Web and it is very hard to know if they are good ontologies or not. For these reasons, it is essential  to evaluate the quality of the ontology after constructing it or before reusing it. 

There are many studies in the literature concerning how to evaluate ontologies \cite{r24}. However, they have some limitations: i) the studies usually present only a set of criteria and questions, but no guidelines to evaluate the ontology; ii) the effort to evaluate an ontology is very high; iii) there is a huge dependence on the expertise of the evaluator to fully understand the criteria and questions; iv) the evaluation is still very subjective.

This work presents a novel methodology for ontology evaluation, called FOCA. The methodology takes into account three fundamental principles: i) it is based on the Goal, Question, Metric (GQM) approach for empirical evaluation (\cite{r17}); ii) the goals of the methodologies are based on the five roles of Knowledge Representation according to (\cite{r3}) and the metrics are based on evaluation criteria proposed by \cite{r9} ; iii) each ontology is evaluated according to the type of ontology, defined by \cite{r10}. FOCA has three main steps: Firstly, the evaluator has to define the type of ontology (i.e. Domain Ontology, Task Ontology of Application Ontology) he/she is interested in evaluating; secondly, he/she iteratively performs the GQM approach; thirdly, the quality (i.e. Partial Quality or Total Quality) of the ontology is calculated. The methodology was evaluated using six ontologists and four ontologies of the same domain (Lattes Domain) created with different ontology methodologies.

The main contributions of this work are: i) to define a step-by-step approach to evaluate the quality of an ontology; ii) to propose an evaluation based on the roles of knowledge representations;
iii) to explicitly differentiate the evaluation according to the type of ontology iii) to have a questionnaire to evaluate the ontologies and iv) to construct a statistical model that automatically assesses the quality of the ontologies.

\section{Background}

In order to better understand the paper, the main types of ontologies are presented; the state of the art on ontology evaluation; the roles of knowledge evaluation and the GQM approach.

\subsection{Types of Ontologies}

Ontologies can be classified into different types, according to the level of abstraction of the universe of discourse the ontology models. According to \cite{r10}, there are four types of ontologies:

\begin{itemize}
	\item Top-level Ontologies: They describe very general concepts such as space, time, love, happiness, etc., which are independent of a particular problem or domain;\\
	
	\item Domain Ontologies and Task Ontologies: They describe general domains, such as medicine, cars, biology or a generic task or activity, such as diagnosing (for medicine) and selling (for cars). These types of ontologies specialize in the concepts introduced in the top-level ontology;\\
	
	\item Application Ontologies: They describe concepts depending on a particular domain and task, which are often specializations of both the related ontologies, such as the medicine course at the Federal University of Alagoas.
\end{itemize}

\subsection{State of the art on Ontology Evaluation}

\cite{r7} proposed a set of criteria to guide the ontology development. The criteria presented by Gruber are: Clarity, Coherence, Extensibility, Minimal encoding bias, Minimal ontological commitment (See Table 1).

\begin{table}[h]
	
	\caption{Quality criteria proposed by Thomas Gruber}
	
	\begin{tabular*}{\columnwidth}{lp{11cm}}
		\hline
		\textbf{Quality criteria}        & \textbf{Description}       \\
		\hline
		Clarity  &  An ontology should effectively communicate the intended meaning of
		defined terms. Definitions should be objective. While the motivation for defining a concept might arise from social situations or computational requirements, the
		definition should be independent of social or computational contexts.\\
		
		Coherence & An ontology should be coherent: that is, it should sanction inferences
		that are consistent with the definitions. At the least, the defining axioms should be
		logically consistent. Coherence should also apply to the concepts that are defined
		informally, such as those described in natural language documentation and
		examples.\\
		
		Extendibility & An ontology should be designed to anticipate the uses of the shared
		vocabulary. It should offer a conceptual foundation for a range of anticipated tasks,
		and the representation should be crafted so that one can extend and specialize the
		ontology monotonically.\\
		
		Minimal encoding bias & The conceptualization should be specified at the
		knowledge level without depending on a particular symbol-level encoding. An
		encoding bias results when a representation choices are made purely for the
		convenience of notation or implementation.\\
		
		Minimal ontological commitment &  An ontology should require the minimal
		ontological commitment sufficient to support the intended knowledge sharing
		activities. An ontology should make as few claims as possible about the world
		being modeled, allowing the parties committed to the ontology freedom to
		specialize and instantiate the ontology as needed.\\
		\hline
	\end{tabular*}
\end{table}

\clearpage

\cite{r8} states that the evaluation process is to determine what the ontology defines correctly, what it does not define and what it defines incorrectly. In addition, his work establishes a set of criteria to evaluate ontologies, as described in Table 2.

\begin{table}[h]
	
	\caption{Quality criteria proposed by G\'omez P\'erez}
	\begin{tabular*}{\columnwidth}{lp{13.5cm}}
		\hline
		\textbf{Quality criteria}        & \textbf{Description}       \\
		\hline
		Consistency  &  This refers to whether it is possible to obtain contradictory
		conclusions from valid input definitions. A given definition is consistent if and
		only if the individual definition is consistent and no contradictory sentences can
		be inferred from other definitions and axioms. .\\
		
		Completeness & Incompleteness is a fundamental problem in ontologies, even more when ontologies are available in such  an open environment as the Semantic Web. In fact, we cannot prove either the completeness of an ontology or the completeness of its definitions, but we can prove both the incompleteness of an individual definition, and, thus, deduce the incompleteness of an ontology, and the incompleteness of an ontology if at least one definition is missing in the established reference framework.\\
		
		Conciseness & An ontology is concise: (a) if it does not store any unnecessary
		or useless definitions, (b) if explicit redundancies between definitions of terms do
		not exist, and (c) if redundancies cannot be inferred from other definitions and
		axioms.\\
		
		Expandability & This refers to the effort required to add new definitions to an
		ontology and more knowledge to its definitions without altering the set of well defined
		properties already guaranteed.\\
		
		Sensitiveness & This relates to how small changes in a definition alter the set of well-defined
		properties already guaranteed.\\
		
		\hline
	\end{tabular*}
\end{table}

\cite{r6} proposed an approach to evaluate and validate ontologies. In his work, Gangemi shows different dimensions for ontology evaluation, such as structural, functional and usability. Table 3 gives details of the criteria (related to dimensions) and their descriptions.

\begin{table}[h]
	
	\caption{Quality criteria proposed by Gangemi}
	
	\begin{tabular*}{\columnwidth}{lp{13cm}}
		\hline
		\textbf{Quality criteria}        & \textbf{Description}       \\
		\hline
		Agreement  &  Measured through the proportion of agreement that experts have with respect to ontology elements, that is, by measuring the consensus of a group of experts.\\
		
		User Satisfaction & This can be evaluated by dedicated research or reliability assessment.\\
		
		Task & This deals with measuring an ontology according to its fitness to some goals, preconditions, postconditions, constraints, options and others.\\
		
		Topic & This measures the ontology according to its fitness for a repository of existing knowledge. \\
		
		Modularity & Modularity measures fitness to a repository of existing reusable components. \\
		\hline
	\end{tabular*}
\end{table}

\cite{r5} proposed to describe the main problems of evaluating ontologies from various views. Before enunciating the problems, Obrst shows the different ways to evaluate an ontology, by (i) a set of ontology criteria, (ii) questions with philosophical perspective and (iii) the verification and validation approach. Table 4 shows the ways of evaluation proposed by Obrst.

\begin{table}[h]
	
	\caption{Ways of evaluation proposed by Leo Obrst}
	
	\begin{tabular*}{\columnwidth}{lp{10cm}}
		\hline
		\textbf{Way of evaluation}        & \textbf{Description}       \\
		\hline
		Quality criteria  & Evaluated against many criteria: Coverage, Complexity, granularity,  Specific Use Cases,  Scenarios, Requirements, Applications, Consistency and Completeness.\\
		
		Questions with philosophical foundations & Is the ontology mappable to some specific upper ontology, so that its evaluation will be at least also partially dependent on the evaluation of the latter? What is the ontology's underlying philosophical theory about reality? (idealist, realist, dimensionalist, etc).
\\
		
		Verification and Validation & Structural, functional and
		usability issues.\\
		\hline
	\end{tabular*}
\end{table}

\cite{r9} collected similar and more common quality criteria from studies carried out by Obrst, Gangemi, Gruber and G\'omez P\'erez . Based on these studies, he defined a set of eight ontology quality criteria to evaluate ontologies. Each quality criteria is organized into questions the evaluator should answer about the ontology, as presented in Table 5.

\begin{table}[h]
	
	\caption{Quality criteria proposed by Denny Vrande\v{c}i\'c}
	
	\begin{tabular*}{\columnwidth}{lp{12cm}}
		\hline
		\textbf{Quality criteria}        & \textbf{Questions}       \\
		\hline
		Accuracy  & Do the axioms comply to the expertise of one or more users? Does the ontology capture and represent correctly aspects of the real world?\\
		
		Adaptability & Does the ontology anticipate its uses? Does it offer a conceptual foundation for a range of anticipated tasks? Can the ontology be extended and specialized monotonically, i.e. without the need to remove axioms? How does the ontology react to small changes in the axioms? Does the ontology comply to procedures for extension, integration, and adaptation?\\
		
		Clarity & Does the ontology communicate effectively the intended meaning of the defined terms? Are the definitions objective and independent of context? Does the ontology use definitions or partial descriptions? Are the definitions documented? Is the ontology understandable?\\
		
		Completeness/Competency & Is the domain of interest appropriately covered? Are competency questions defined? Can the ontology answer them? Does the ontology include all relevant concepts and their lexical representations?\\
		
		Computational efficiency & How easy and successful can reasoners
		process the ontology? How fast can the usual reasoning services (satisfiability, instance classification, querying, etc.) be applied to the ontology?\\
		
		Conciseness & Does the ontology include irrelevant axioms with regards to the domain to be covered (i.e. a book ontology including axioms about African lions)? Does it include redundant axioms? Does it impose a minimal ontological commitment, i.e. does it specifying the weakest theory possible and define only essential terms? How weak are the assumptions regarding the ontology's underlying philosophical theory about reality?\\
		
		Consistency/Coherence & Do the axioms lead to contradictions (logical consistency)? Are the formal and informal descriptions of the ontology consistent, i.e. does the documentation match the specification? Does the translation from the knowledge level to the encoding show a minimal encoding bias? Are any representation choices made purely for the convenience of notation or implementation.\\
		
		Organizational fitness &  Is the ontology easily deployed within the organization? Do ontology-based tools within the organization put constraints upon the ontology? Was the proper process for creating the ontology used? Was it certified, if required? Does it meet legal requirements? Is it easy to access? Does it align to other ontologies already in use? Is it well shared among potential stakeholders?\\
		
		\hline
	\end{tabular*}
\end{table}

\cite{r24} used the same metrics proposed by the aforementioned authors to elaborate a categorization of measures for ontology evaluation. He presented two perspectives of evaluation: Ontology Correctness and Ontology Quality. For each perspective, he made a correspondence between the metrics proposed by the authors and for each metric, their measures. Table 6 shows the correspondence between the perspectives, metrics and measures.

\begin{table*}[ht]
	\centering
	\caption{The evaluation perspectives with the existing metrics and their measures}
	\begin{tabular}    {p{0.25\linewidth}p{0.17\linewidth}p{0.50\linewidth}}
		\hline
		\textbf{Evaluation Perspective} & 	\textbf{Metric} & 	\textbf{Measure}\\
		\hline
		Ontology Correctness & Accuracy &  precision: total number correctly found over whole knowledge defined in ontology \\
		&  & Recall: total number correctly found over whole knowledge that should be found \\
		&  & Coverage \\
		& Completeness & Coverage \\
		& Conciseness &  \\
		& Consistency & Count: Number of terms with inconsistent meaning \\
		Ontology Quality & Computational efficiency &  Size \\
		& Adaptability & Coupling: Number of external classes referenced \\
		&  & Cohesion: Number of Root (NoR), Number of Leaf (NoL), Average \\
		&  & Depth of Inheritance Tree of Leaf Nodes(ADIT-LN) \\
		& Clarity & Number of word senses \\
		
		\hline
	\end{tabular}
	
\end{table*}

There are various studies about ontology evaluation using different criteria to analyze the modeling quality, the speed of computational services, the meaning of the concepts and other perspectives. Thus, each study presents criteria and its definitions, but these studies make this task very difficult to perform. In addition, there is a dependence on the expertise to understand the actual criteria and questions to verify them. Moreover, \cite{r9} says that the lack of such experimental evaluations to match methods and criteria will hinder meaningful ontology evaluations.

\subsection{Roles of Knowledge Representation}

The goal of this section is to present the roles of knowledge representation proposed by \cite{r3}. The authors presented five roles in order to describe what is knowledge representation. Due to its clear definition, we used these roles to describe FOCA methodology. It is important to say that the use of the proposed roles of knowledge representation with the questions and metrics proposed in the FOCA methodology support the features of the ontological level (i.e. \cite{r21}). The five roles of KR are described as follows:

\begin{itemize}
	
	\item Substitute. This role represents how the ontology approaches the real world, i.e., which concepts were captured and which were omitted. For example, the bicycle's concepts are vehicle, tires, handlebars, rim.  The closer the concepts are to the real world, the more this role will be fulfilled;\\
	
	\item Ontological Commitments. This role represents how the ontology is closer to the real world, i.e., the more consistent the representation is, the better this role will be fulfilled. For example, the representation \textit{bicycle is a vehicle, vehicle is an object} is more consistent than \textit{bicycle is an object} or \textit{the bicycle has two tires, and not one or three or more};\\
	
	\item Intelligent Reasoning. This role represents how the ontology correctly infers the real world, i.e., the more correct the relations and attributes are defined, the more this role will be fulfilled. For example,  \textit{there is a vehicle which has two tires, handlebars and thin wheels. Therefore, this vehicle is a bicycle};\\
	
	\item Efficient Computation. This role represents how the machine can think about a domain in computer applications . For example, if a user is looking for a bicycle which is blue, has two big tires and manufactured by x, if all the web sites have the bicycle ontology, the machine can find this bicycle in a few seconds;\\
	
	\item Human Expression. This role represents how easy it is to understand the modeling, i.e., the clearer the concepts and their relations, the more this role will be fulfilled. For example, the bicycle is represented by \textit{bicycle}, and not  \textit{bi} or  \textit{bic}.\\
\end{itemize}

\subsection{The Goal-Question-Metric Approach}

According to \cite{r17}, any engineering process requires feedback and evaluation. Measures can be taken by developers, managers, customers and the corporation in order to help understand and control the artifact processes and product, and the relationships between them. Thus, there are various objectives (with different verifications) defined by these actors, aiming at measuring the quality of the artifact.

For an organization to measure properly, the objectives should be specified and also how these goals will be checked. The Goal/Question/Metric (GQM) approach supports these steps. The goals are defined and refined in a set of questions which are used to extract information from the models. Similarly, the questions define a set of metrics and data to collect and provide a framework for interpretation. Figure 2 shows the GQM approach.

\clearpage

\begin{figure}[!htb]
	\centering
	\includegraphics{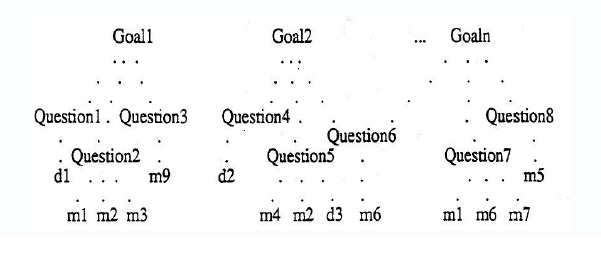}
	\caption{The GQM Approach, proposed by \cite{r17}}
	\label{}
\end{figure}

\section{The FOCA Methodology}

This section presents the main contribution of this paper. Firstly, this section presents the main objectives of the methodology. Secondly, how the methodology was developed will be described. Afterwards, the steps of the methodology will be presented, and finally how to use the methodology will be shown. 

There are several problems that an ontologist could encounter when trying to model an ontology, such as the difficulty of choosing an appropriate methodology. According to the complexity of the domain, respecting the knowledge representation roles are reasons that currently motivate an ontologist to evaluate the quality of the ontology, but current approaches which assess ontologies do not have guidelines to evaluate the ontology step-by-step and much effort is required. In addition, there is a huge dependence on the evaluator's expertise to understand the criteria and questions in-depth, and also the evaluation is still very subjective.

The FOCA Methodology takes into account the main quality criteria proposed by \cite{r9}. The quality criteria are matched with the five roles of KR according to similarity, using the GQM approach. This approach was chosen because there is a similarity between general aspects of evaluation (corresponding goals and roles), specific aspects (corresponding metrics and criteria) and their questions. 

The roles of knowledge representation, which are the objectives of the evaluation, were considered as GOALs according to the GQM approach. The quality criteria, which are the measurement units, were considered as the METRICs of the GQM approach. Finally, the questions that are concerned with the metrics were considered as QUESTIONs of the GQM approach. Table 7 shows the reasons of justification between the Questions and Roles of KR.

\clearpage

\begin{table*}[ht]
	\centering
	\caption{The GQM of FOCA Methodology}
	\begin{tabular}    {p{0.23\linewidth}p{0.07\linewidth}p{0.63\linewidth}}
		\hline
		\textbf{Correspondent Role} & 	\textbf{Question} & 	\textbf{Justification}\\
		\hline
		
		Substitute & Q1 & This question addresses the document which contains the competency questions, the main terms and the objectives of the ontology.

		\\
		
		Substitute & Q2 & This question addresses the verification of the coherence between the documentation which contains the competency questions, the main terms and the objectives of the ontology with the model.
		\\
		Substitute & Q3 & This question addresses reusing existing concepts to model the real
world.
		 \\
		
		Ontological Commitments & Q4 & This question addresses choosing the best representation for a  specific domain.
		 \\
		
		Ontological Commitments & Q4 & This question addresses choosing the best representation for a more abstract domain.
 		\\
		
		Ontological Commitments & Q6 & This question addresses verifying the coherence with the real world.
		 \\
		
		Intelligent Reasoning & Q7 & This question addresses verifying a correct reasoning of ontology
 		\\
		
		Intelligent Reasoning & Q8 & This question addresses verifying a correct reasoning of ontology. 
		\\
		
		Efficient Computation & Q9 & This question addresses a good computational performance (successful). 
		\\
		
		Efficient Computation & Q10 & This question addresses a good computational performance (speed).
		\\
		
		Human Expression & Q11 & addresses the easy understanding of the modeling.
		 \\
		
		Human Expression & Q12 & This question addresses the easy understanding of modeling.
		\\
		
		Human Expression & Q13 & This question addresses the easy understanding of the modeling. 
		\\
		
		\hline
	\end{tabular}
	
\end{table*}

After corresponding the questions with the roles, the GQM structure for the FOCA Methodology was created. Table 8 shows the roles organized into GOALs, the questions organized into QUESTIONs and the criteria organized into METRICs.

\begin{table*}[ht]
	\centering
	\caption{The GQM of FOCA Methodology}
	\begin{tabular}    {p{0.20\linewidth}p{0.50\linewidth}p{0.20\linewidth}}
		\hline
		\textbf{Goal} & 	\textbf{Question} & 	\textbf{Metric}\\
		\hline
		1. Check if the ontology complies with Substitute. & Q1. Were the competency questions defined? & 1. Completeness.
		\\
		& Q2. Were the competency questions answered? & 1. Completeness.\\
		& Q3. Did the ontology reuse other ontologies? & 2. Adaptability. \\
		2. Check if the ontology complies Ontological Commitments.& Q4. Did the ontology impose a minimal ontological commitment? & 3. Conciseness.\\
		& Q5. Did the ontology impose a maximum ontological commitment? & 3. Conciseness. \\
		& Q6. Are the ontology properties coherent with the domain? & 4. Consistency. \\
		3. Check if the ontology complies with Intelligent Reasoning & Q7. Are there contradictory axioms? & 4. Consistency. \\
		& Q8. Are there redundant axioms? & 3. Conciseness. \\
		4. Check if the ontology complies Efficient Computation & Q9. Did the reasoner bring modelling errors? & 5. Computational efficiency. \\
		& Q10. Did the reasoner perform quickly? & 5. Computational efficiency. \\
		5. Check if the ontology complies with Human Expression.& Q11. Is the documentation consistent with modelling? & 6. Clarity. \\
		& Q12. Were the concepts well written? & 6. Clarity. \\
		& Q13. Are there annotations in the ontology that show the definitions of the concepts? & 6. Clarity. \\

		\hline
	\end{tabular}
	
\end{table*}

Figures (3 and 4) illustrate two correspondences in a GQM format.

\clearpage 

\begin{figure}[!htb]
	\centering
	\includegraphics[scale=0.6]{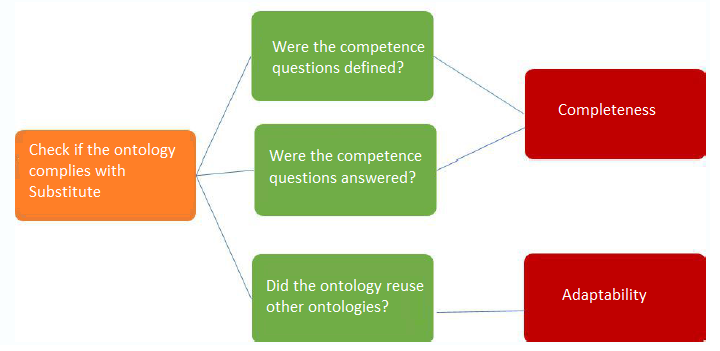}
	\caption{Example of GOAL 1 in GQM format}
	\label{}
\end{figure}

\begin{figure}[!htb]
	\centering
	\includegraphics[scale=0.6]{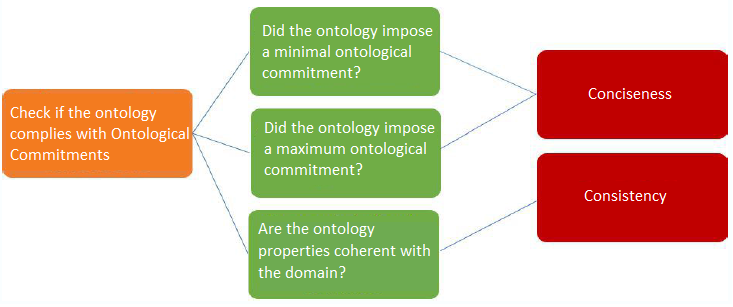}
	\caption{Example of GOAL 2 in GQM format}
	\label{}
\end{figure}

With regards the methodology it consists of three steps: 

\begin{enumerate}
	\item \textbf{Ontology Type Verification}: An ontology does not need to run all the questions, because there are contradictory questions, for example, questions 4 and 5. These questions are only verified according to the ontology type (see subsection 2.1). Thus, the evaluator should verify if the ontology is:
	
	\begin{itemize}
		\item \textbf{Type 1: A Domain or Task ontology}. The ontology describes, respectively, the vocabulary of a generic domain (such as medicine or vehicles) or a generic task (such as selling or diagnosing). If the ontology is Type 1, question 4 should not be verified, because it only asks for specific domains; or\\
		
		\item \textbf{Type 2: Application ontology}. This ontology describes concepts that depend on a very specific domain (student registration system of a medicine course of a specific university, for example). If the ontology is Type 2, question 5 should not be verified.\\
		
	\end{itemize}
	
	It is important to note that TOP ONTOLOGY was not considered, because the questions of this methodology are not enough to evaluate the level of abstraction that this ontology type requires.\\
	
	\item \textbf{Questions Verification}: In this step, the evaluator will answer the questions following the GQM approach. After answering the questions, the evaluator will establish a grade\footnote{The grades will be explained after.} for each question. The questions of this step were described in Table 8.
	
	\begin{itemize}
		\item Q1. Were the competency questions defined? 
		\item Q2. Were the competency questions answered?
		\item Q3. Did the ontology reuse other ontologies?
		\item Q4. Did the ontology impose a minimal ontological commitment?
		\item Q5. Did the ontology impose a maximum ontological commitment?
		\item Q6. Are the ontology properties coherent with the domain? 
		\item Q7. Are there contradictory axioms?
		\item Q8. Are there redundant axioms?
		\item Q9. Does the reasoner bring modelling errors?
		\item Q10. Does the reasoner perform quickly?
		\item Q11. Is the documentation consistent with the modelling?
		\item Q12. Were the concepts well written?
		\item Q13. Are there annotations in the ontology bringing the concepts definitions?
	\end{itemize}
		
	\item \textbf{Quality Verification}: In this step, the evaluator verifies all the questions from Step 2 and calculates the correspondent grades. After that, he/she must calculate the quality of the ontology. The quality of the ontology (a score in (0,1)) is calculated by the beta regression models, proposed by \cite{r15}. \footnote{The equation was validated and presented in the next section.}\footnote{This step can be automated.}
	
	\begin{center}
	\begin{scriptsize}  
		$$
		\hskip-0.0truein		\widehat\mu_i =
		\frac{\exp \{-0.44+ 0.03{(Cov_S \times Sb)}_i + 0.02 {(Cov_C \times Co)}_i + 0.01{(Cov_R \times Re)}_i + 0.02{(Cov_{Cp} \times Cp)}_i - 0.66 LExp_i - 25{(0.1 \times Nl)}_i\}}
		{1+ {\exp \{-0.44+ 0.03{(Cov_S \times Sb)}_i + 0.02 {(Cov_C \times Co)}_i + 0.01{(Cov_R \times Re)}_i + 0.02{(Cov_{Cp} \times Cp)}_i - 0.66 LExp_i - 25{(0.1 \times Nl)}_i\}}}
		$$
	\end{scriptsize}
	\end{center}
	
	The quality of the ontology can be calculated in two ways:
	
	\begin{itemize}
		\item \textbf{Total Quality}: Evaluator "i" calculates the quality of the ontology considering the 5 roles of KR. To calculate the total quality:
		\begin{itemize}		 
			\item \textbf{$Cov_S$} is the mean of grades obtained from Goal 1. Note that Question 1 contains three sub-questions, therefore the grade of Question 1 is the mean between the three sub-questions. Finally, $Cov_S$ is the mean between Question 1, Question 2 and Question 3;
			\item \textbf{$Cov_C$} is the mean of grades obtained from Goal 2. Note that the mean will be between Question 5 and Question 6 if the ontology is Type 1 and the mean will be between Question 4 and Question 6 if the ontology is Type 2;
			\item \textbf{$Cov_R$} is the mean of grades obtained from Goal 3;
			\item \textbf{$Cov_Cp$} is the mean of grades obtained from Goal 4;
			\item \textbf{$LExp$} is the variable which corresponds with the experience of the evaluator. If the evaluator considers himself/herself a person with vast experience in ontologies, the value of $LExp$ is 1, if not, 0; 
			\item \textbf{$Nl$} is 1 only if some Goal was impossible for the evaluator to answer all the questions;
			\item \textbf{$Sb=1$}, \textbf{$Co=1$}, \textbf{$Re=1$}, \textbf{$Cp=1$}, because the total quality considers all the roles.
		\end{itemize}
		\item \textbf{Partial Quality}: Evaluator "i" calculates the quality of the ontology considering only some roles (for example: to consider only Substitute and Intelligent Reasoning). To calculate the partial quality:
		\begin{itemize}
			\item If the evaluator wants to evaluate the ontology \textbf{in terms of Substitute}, he/she will consider:
			\begin{itemize}
				\item $Cov_S$, the mean of grades obtained from Goal 1;
				\item $Sb=1$, because the evaluator is considering only Goal 1;
				\item $Cov_C=0$, $Cov_R=0$, $Cov_Cp=0$, $Co=0$, $Re=0$, $Cp=0$, because the evaluator does not consider Goals 2,3,4,5;
				\item Give the values to $LExp$ and $Nl$.
			\end{itemize}
			\item If the evaluator wants to evaluate the ontology \textbf{in terms of Ontological Commitments}, he/she will consider:
			\begin{itemize}
				\item $Cov_C$, the mean of grades obtained from Goal 2;
				\item $Co=1$, because the evaluator is considering only Goal 2;
				\item $Cov_S=0$, $Cov_R=0$, $Cov_Cp=0$, $Sb=0$, $Re=0$, $Cp=0$, because the evaluator does not consider Goals 1,3,4,5;
				\item Give the values to $LExp$ and $Nl$.
			\end{itemize}
			\item If the evaluator wants to evaluate the ontology \textbf{in terms of Intelligent Reasoning}, he/she will consider:
			\begin{itemize}
				\item $Cov_R$, the mean of grades obtained from Goal 3;
				\item $Re=1$, because the evaluator is considering only Goal 3;
				\item $Cov_S=0$, $Cov_C=0$, $Cov_Cp=0$, $Sb=0$, $Co=0$, $Cp=0$, because the evaluator does not consider Goals 1,2,4,5;
				\item Give the values to $LExp$ and $Nl$.
			\end{itemize}
			\item If the evaluator wants to evaluate the ontology \textbf{in terms of Efficient Computation}, he/she will consider:
			\begin{itemize}
				\item $Cov_Cp$, the mean of grades obtained from Goal 4;
				\item $Cp=1$, because the evaluator is considering only Goal 4;
				\item $Cov_S=0$, $Cov_C=0$, $Cov_R=0$, $Sb=0$, $Co=0$, $R=0$, because the evaluator does not consider Goals 1,2,3,5;
				\item Give the values to $LExp$ and $Nl$.
			\end{itemize}
			\item If the evaluator wants to evaluate the ontology \textbf{in terms of Human Expression}, he/she will consider:
			\begin{itemize}
				\item $Cov_S=0$, $Cov_C=0$, $Cov_R=0$, $Cov_Cp=0$, $Sb=0$, $Co=0$, $R=0$, $Cp=0$ because the evaluator does not consider Goals 1,2,3,4;
				\item Give the values to $LExp$ and $Nl$.
			\end{itemize}
		\end{itemize}
	\end{itemize}
	
	For example, if the evaluator wants to evaluate the ontology in terms of Substitute and Efficient Computation, the equation is:
	
	\begin{scriptsize}  
		$$
		\hskip-0.0truein		\widehat\mu_i =
		\frac{\exp \{-0.44+ 0.03{(Cov_S \times Sb)}_i + 0.02{(Cov_{Cp} \times Cp)}_i - 0.66 LExp_i - 25{(0.1 \times Nl)}_i\}}
		{1+ {\exp \{-0.44+ 0.03{(Cov_S \times Sb)}_i + 0.02{(Cov_{Cp} \times Cp)}_i - 0.66 LExp_i - 25{(0.1 \times Nl)}_i\}}}
		$$
	\end{scriptsize}\\
\end{enumerate}

Figure 5 resumes the methodology.

\clearpage

\begin{figure}[!htb]
	\centering
	\includegraphics[scale=0.7]{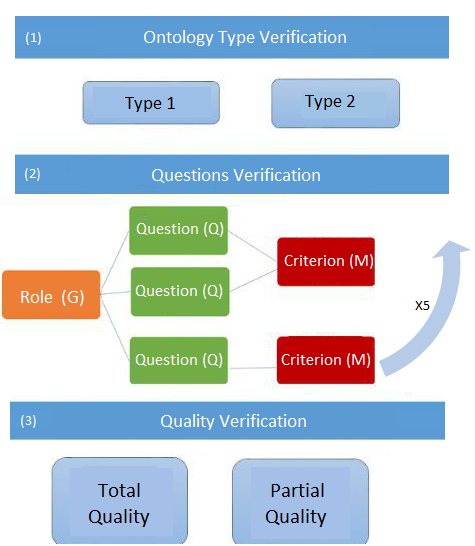}
	\caption{The FOCA methodology steps}
	\label{}
\end{figure}

\subsection{Execution the Methodology}

Initially, to execute the methodology the evaluator needs to know what type of ontology will be evaluated (Step 1). To do this, the evaluator reflects on the universe of discourse that is represented by the ontology. If the ontology models an abstract, a subject or knowledge area, the ontology is classified as  \textbf{Type 1} (Task or Domain ontology) and if the ontology models a concept, which specifies or instantiates an abstract concept, subject or knowledge area for a particular domain, the ontology is classified as \textbf{Type 2} (Application ontology). For example:

\begin{itemize}
	\item The FOAF\footnote{http://xmlns.com/foaf/spec/} ontology models relations between people, i.e., it uses the people concept and specializes. Thus, the FOAF ontology can be classified as Type 2;
	\item The Locality\footnote{http://linkn.com.br/linkn/onto/locality/} ontology models the geographic limitations and locations, i.e., primitive and  abstract concepts. Thus, the Locality ontology can be classified as Type 1.

\end{itemize}

After the ontology type verification, the evaluator can already verify the questions following the GQM approach (Step 2). To verify the model, the evaluator has 12 questions to answer, because question 4 or question 5 will not be evaluated. To do this, he/she must follow the GQM approach and may evaluate it in any order of goals he/she wishes. The next tables show how each question should be verified.

\begin{table*}[ht]
	\centering
	\caption{How to verify the Questions of Goal 1}
	\begin{tabular}    {p{0.05\linewidth}p{0.08\linewidth}p{0.80\linewidth}}
		\hline
		\textbf{Goal} & 	\textbf{Question} & 	\textbf{How to Verify}\\
		\hline
		1 & Q1 & Firstly, check if the document has the ontology competencies defined. If they do not exist, the grade is
0. If they exist, answer three sub-questions: Does the document define the ontology objective? (for example: "This ontology models the domain of..."); Does the document define the ontology stakeholders?
(for example: "This ontology should be used by..."); Does the document define the use of scenarios?
(i.e., the situations in which the ontology must be used). For each sub-question, give one of these grades:
25,50,75,100. Finally, the mean of the three sub-questions must be calculated.
 \\
		\hline
		1 & Q2 & If you established grade 0 in the previous question, the competencies were not defined and you cannot
evaluate this question. Thus, the grade of this question is 0. If the competencies exist, check if the
ontology responds to what was defined in the competencies document. Grades: 25,50,75,100.
\\
		\hline
		1 & Q3 & Check if the ontology reuses other ontologies. If it does not, the grade is 0. If it does, the grade is 100.\\
		\hline
	\end{tabular}
\end{table*}

\begin{table*}[ht]
	\centering
	\caption{How to verify the Questions of Goal 2}
	\begin{tabular}    {p{0.05\linewidth}p{0.08\linewidth}p{0.80\linewidth}}
		\hline
		\textbf{Goal} & 	\textbf{Question} & 	\textbf{How to Verify}\\
		\hline
		2 & Q4 & \textbf{This question should only be checked if the ontology is type 2. If the ontology is type 1, go to the next question}.In this question, check if the ontology does not use much abstraction to define the concepts. If the ontology is full of abstraction (for example: an ontology which models the facebook site does not need to define what a computer system is, or what a computer is, and other abstraction concepts), the grade is 0. If there are only some abstractions, give a grade between these: 25 (very specific), 50 (moderate abstraction), 75 (many abstractions), 100 (full of abstractions).\\
		\hline
		2 & Q5 & \textbf{This question should only be checked if the ontology is type 1}. In this question, check if the ontology uses primitive concepts to define the evaluated domain (for example, an ontology which models a person, uses the concepts \textit{thing $\to$ living being $\to$ human being $\to$ person} to define the person concept). If the ontology does not use abstractions, the grade is 0. If there is only some abstractions, give a grade between these: 25 (very specific), 50 (moderate abstraction), 75 (much abstractions), 100 (full of abstractions).\\
		\hline
		2 & Q6 & In this question, check if the classes and properties are coherent with the modelled domain. If the ontology is full of incoherences (for example, an ontology which models the concept car has a class lion and the property quantityOfPaws, that is, do not exist in domain), the grade is 0. If there are some incoherences, give a grade between these: 25,50,75. If there is no incoherence, the grade is 100.\\
		\hline
	\end{tabular}
\end{table*}

\begin{table*}[ht]
	\centering
	\caption{How to verify the Questions of Goal 3}
	\begin{tabular}    {p{0.05\linewidth}p{0.08\linewidth}p{0.80\linewidth}}
		\hline
		\textbf{Goal} & 	\textbf{Question} & 	\textbf{How to Verify}\\
		\hline
		3 & Q7 & In this question, check if the classes and properties (functional, transitive, reflexive and others) characteristics contradict the domain (for example: \textit{LivingBeing} is a subclass of \textit{Person} in an ontology which models the person concept or \textit{socialSecurityNumber} is not a \textit{functional} property, because a person cannot have more than one Social Security Number). If the ontology is full of contradictions, the grade is 0. If there are some contradictions, give a grade between these: 25,50,75. If there are no contradictions, the grade is 100.\\
		\hline
		3 & Q8 & In this question, check if there are classes or properties which model the same thing with the same meaning (for example, use mouse for hardware and animal). If the ontology is full of redundancies, the grade is 0. If there are some redundancies, give a grade between these: 25,50,75. If there are no contradictions, the grade is 100.\\
		\hline
	\end{tabular}
\end{table*}

\begin{table*}[ht]
	\centering
	\caption{How to verify the Questions of Goal 4}
	\begin{tabular}    {p{0.05\linewidth}p{0.08\linewidth}p{0.80\linewidth}}
		\hline
		\textbf{Goal} & 	\textbf{Question} & 	\textbf{How to Verify}\\
		\hline
		4 & Q9 & \textbf{Save all your records here}. In this question, check if, running the reasoner, returns some kind of error. If the ontology is full of errors (or the software stops responding), the grade is 0. If there are some errors, give a grade between these: 25,50,75. If there are no errors, the grade is 100.\\
		\hline
		4 & Q10 & In this question, check if the reasoner is running quickly. If the reasoner stops, the grade is 0. If there is any delay, give a grade between these: 25,50,75. If it runs quickly, the grade is 100.\\
		\hline
	\end{tabular}
\end{table*}

\clearpage

\begin{table*}[ht]
	\centering
	\caption{How to verify the Questions of Goal 5}
	\begin{tabular}    {p{0.05\linewidth}p{0.08\linewidth}p{0.80\linewidth}}
		\hline
		\textbf{Goal} & 	\textbf{Question} & 	\textbf{How to Verify}\\
		\hline
		5 & Q11 & In this question, check if the documentation of ontology exists. If it does not exist, the grade is 0. If the
documentation exists, answer two sub-questions: Are the written terms in the documentation the same as the modelling?; Does the documentation explain what each term is and does it justify each detail of modeling?  For each sub-question, give one of these grades: 25,50,75,100. Finally, the mean of two sub-questions must be calculated.\\
		\hline
		5 & Q12 & In this question, check if the classes or properties of ontology are written in an understandable and correct form (according to English or another language). If the ontology is difficult to understand or full of poorly written terms, the grade is 0. If there are some errors or a mix of languages, give the grade
between these: 25,50,75. If the ontology is well written and one language was used, 100.
\\
		\hline
		5 & Q13 & In this question, check if the existing annotations bring the definitions of the modelled concepts. If there are no annotations, the grade is 0. If there are some annotations, give a grade between these: 25,50,75. If all the concepts have annotations, the grade is 100.
\\
		\hline
	\end{tabular}
\end{table*}

After the evaluator evaluates each question and gives the correspondent grades, he/she is ready to calculate the number that represents the ontology quality (Step 3), i.e., he/she puts the numerical values in the equation presented previously. For example, suppose the evaluator has given the follow grades presented in the following Table:

\begin{table*}[ht]
	\centering
	\caption{Example of grades given by the evaluator}
	\begin{tabular}{p{0.08\linewidth}p{0.08\linewidth}}
		\hline
		\textbf{Question} & \textbf{Grade}\\
		\hline
		Q1 & 50 \\
		Q2 & 75\\
		Q3 & 100\\
		Q4 & 25\\
		Q5 & -\\
		Q6 & 50\\
		Q7 & 25\\
		Q8 & 50\\
		Q9 & 100\\
		Q10 & 100\\
		Q11 & 75\\
		Q12 & 75\\
		Q13 & 25\\
		\hline
	\end{tabular}
\end{table*}

Now, the evaluator must calculate the mean of each Goal. The next Table shows the mean of each Goal from the grades given in the example:

\begin{table*}[ht]
	\centering
	\caption{The Goals Means from the grades given by the evaluator}
	\begin{tabular}{p{0.08\linewidth}p{0.08\linewidth}}
		\hline
		\textbf{Goal} & \textbf{Mean}\\
		\hline
		1 & 75 \\
		2 & 37.5\\
		3 & 37.5\\
		4 & 100\\
		5 & 58.33\\
		\hline
	\end{tabular}
\end{table*}

After the evaluator calculates the means, he/she must put the values in the equation. Suppose the evaluator wants to calculate the Total Quality. The equation, using the means presented in the previous table, and considering that the evaluator has vast experience, is:

\begin{center}
	\begin{scriptsize}  
		$$
		\hskip-0.0truein		\widehat\mu =
		\frac{\exp \{-0.44+ 0.03{(75 \times 1)} + 0.02 {(37.5 \times 1)} + 0.01{(37.5 \times 1)} + 0.02{(100 \times 1)} - 0.66 \times 1 - 25{(0.1 \times 0)}\}}
		{1+ {\exp \{-0.44+ 0.03{(75 \times 1)} + 0.02 {(37.5 \times 1)} + 0.01{(37.5 \times 1)} + 0.02{(100 \times 1)} - 0.66 \times 1 - 25{(0.1 \times 0)}\}}}
		$$
	\end{scriptsize}
\end{center}

\begin{center}
	\begin{scriptsize}  
		$$
		\hskip-0.0truein		\widehat\mu = 0.986278841
		$$
	\end{scriptsize}
\end{center}

Thus, the total quality of the ontology is 0.986278841.

Now, suppose the evaluator wants to calculate the quality of the ontology in terms of Goals 2 and 3. The equation, using the means presented in the previous Table, and considering that the evaluator has vast experience, is:

\begin{center}
	\begin{scriptsize}  
		$$
		\hskip-0.0truein		\widehat\mu =
		\frac{\exp \{-0.44+ 0.03{(75 \times 0)} + 0.02 {(37.5 \times 1)} + 0.01{(37.5 \times 1)} + 0.02{(100 \times 0)} - 0.66 \times 1 - 25{(0.1 \times 0)}\}}
		{1+ {\exp \{-0.44+ 0.03{(75 \times 0)} + 0.02 {(37.5 \times 1)} + 0.01{(37.5 \times 1)} + 0.02{(100 \times 0)} - 0.66 \times 1 - 25{(0.1 \times 0)}\}}}
		$$
	\end{scriptsize}
\end{center}

\begin{center}
	\begin{scriptsize}  
		$$
		\hskip-0.0truein		\widehat\mu =
		\frac{\exp \{-0.44+ 0.02 {(37.5 \times 1)} + 0.01{(37.5 \times 1)} - 0.66 \times 1 - 25{(0.1 \times 0)}\}}
		{1+ {\exp \{-0.44+ 0.02 {(37.5 \times 1)} + 0.01{(37.5 \times 1)} - 0.66 \times 1 - 25{(0.1 \times 0)}\}}}
		$$
	\end{scriptsize}
\end{center}

\begin{center}
	\begin{scriptsize}  
		$$
		\hskip-0.0truein		\widehat\mu = 0.506249674
		$$
	\end{scriptsize}
\end{center}

Thus, the partial quality of the ontology, considering the roles of Ontological Commitments and Intelligent Reasoning is 0.506249674.

\section{Definition of the statistical model and Validation of the FOCA Methodology}

In this section, the construction of the statistical model and the FOCA methodology validation are described. First, a study about the definition of a problem was made. Second, research questions were formulated. After that, an empirical study with six people with different levels of expertise in ontology was conducted, where each person evaluated ontologies with only the questions and, afterwards, used the methodology. 

\subsection{Problem definition}

This empirical study is in the context of ontology quality verification answering questions following a step-by-step process. The verification of ontology in this work involves information such as: the type of ontology, its use, its real world representation, well represented concepts, the computational efficiency, easy understanding of modelling and other features. If a methodology does not establish a step-by-step process or explain how to verify each of these features, evaluating the ontologies becomes very difficult and imprecise. Based on this problem, the FOCA Methodology aims to correspond these features with a more general thing, i.e., the types of ontology presented by \cite{r10}, and the five roles of KR presented by \cite{r3}. The correspondence are: 

\begin{itemize}
	\item The type of ontology $\to$ \textit{domain or task} or \textit{application} ontology;
	\item The use of ontology $\to$ competencies, main terms and reuse in \textit{Substitute};
	\item Representing the real world $\to$ minimal and maximum commitments and coherence in \textit{Ontological Commitments};
	\item Well represented concepts $\to$ contradictions and redundancies in \textit{Intelligent Reasoning};
	\item The computational efficiency $\to$ fast execution and successfully of reasoners in \textit{Efficient Computation};
	\item The easy understanding of modelling $\to$ documentation, writing and annotations in \textit{Human Expression}.
\end{itemize}

However, establishing a step-by-step process and corresponding these features leads to some questions, such as "Does the methodology guide the evaluator well?" or "Is the methodology dependent or independent on the evaluator's experience (in ontologies), i.e., can the methodology guide an evaluator who has vast experience and one who is inexperienced to come to the same results?". Thus, the grades and experience are the main metrics for the empirical study. From these questions, metrics and the business problem are presented, and the technical problems entailed should also be considered. To answer the business problem, it is important to evaluate if it is appropriate to make a correspondence between the knowledge representation roles and quality criteria to evaluate the ontology. Thus, the technical problem makes us reflect on some questions: "Does establishing a step-by-step guide minimize subjectivity?" or "Is the methodology appropriate for people with any experience in the area of ontologies?".

\subsection{Research questions}

The objective of the empirical study is to check if the subjective scores (the grades given to people who did not use the methodology) is the same or similar to the objective scores (the grades given to people who used the methodology). All the grades will be used to construct a statistical equation to calculate the quality ontology, total or partial. If it is possible to create these statistical models, the methodology will be validated. Thus, the research questions and hypotheses involve validating the methodology and the possibility of creating the equation. The research questions and hypotheses are:

\begin{itemize}

	\item \textbf{Research Question 1}: Is the methodology appropriate for people with any experience in ontologies?
		\begin{itemize}
			\item \textit{Research Question 1.1}: Can the methodology approximate grades of people with vast and little experience?
			\item \textit{Research Question 1.2}: Can the methodology reduce the effort of people when evaluating ontologies?

		\end{itemize}
		
	\item \textbf{Research Question 2}: Is the methodology valid?
		\begin{itemize}
			\item \textit{Research Question 2.1}: Is it possible to create an equation (statistical model) to calculate the total quality of the ontology?
			\item \textit{Research Question 2.2}: Is it possible to create an equation (statistical model) to calculate the partial
quality of the ontology?
		\end{itemize}

\end{itemize}

\subsection{Conducting the empirical study}

In order to conduct the empirical study, 4 ontologies were implemented, which modelled the Lattes Platform. According to \cite{r11}, the Lattes Platform is the major scientific information system maintained by the National Council for Scientific and Technological Development (CNPq). This platform manages the curricular information of researchers (and institutions) working in Brazil based on the so-called Lattes Curriculum. The ontologies were implemented by groups of four students of ontology disciplines at the Federal University of Alagoas (Computing Institute), Brazil. From the 4 ontologies, 3 were implemented with a methodology for ontology engineering, and 1 was implemented without the methodology. The following table shows the ontologies and their context and corresponding methodologies.

\clearpage

\begin{table}[h]

\caption{Ontologies and Correspondent Methodologies}
\begin{tabular*}{\linewidth}{llp{15cm}}
\hline
\textbf{Ontology} & \textbf{Correspondent Methodology}       \\
\hline
Ontology A & HCOME [\cite{r12}] \\

Ontology B & Gr\"uninger and Fox [\cite{r13}] \\

Ontology C & Methontology [\cite{r14}] \\

Ontology D & Without\\

\hline
\end{tabular*}
\end{table}

After implementing the ontologies, 6 people with different experiences in the ontology area were selected to evaluate each ontology. An experience consists of "how long the person studied", publications in the ontology area and practical experience with modeling. The following table shows the people, called evaluators, and their correspondent experiences.

\begin{table}[h]

\caption{Evaluators and Correspondent Experiences}
\begin{tabular*}{\linewidth}{llp{15cm}}
\hline
\textbf{Evaluator} & \textbf{Correspondent Experience}       \\
\hline
Evaluator 1 & Very large  \\

Evaluator 2 & Very large  \\

Evaluator 3 & Large \\

Evaluator 4 & Medium\\

Evaluator 5 & Low\\

Evaluator 6 & Very low\\

\hline
\end{tabular*}
\end{table}

The empirical study had two phases. In the first phase, each evaluator received a document including 13 questions, without GQM, Roles of Knowledge Representation and "How to verify". The evaluators answered the questions based on their experience, giving grades (called subjective scores) between 0.001 and 0.990. If an evaluator did not know how to answer a question, he/she did not assign any grade. Each evaluator followed the same process for the 4 ontologies.

In the second phase, each evaluator received a document including the FOCA Methodology, i.e., steps 1 and 2 mentioned before. Each evaluator followed the same process for the 4 ontologies, giving the grades (called objective scores) presented in the methodology.

The subject and object definition was based on statistical methods.

\subsection{Results}

To answer the research questions, all the grades were collected from the 6 people. It was observed that the medians of the subjective scores of more experienced evaluators were lower than the medians of the evaluators with less experience, because the evaluators with more experience were stricter when evaluating. The following figure shows the medians.

\begin{figure}[!htb]
\centering
\includegraphics[scale=0.5]{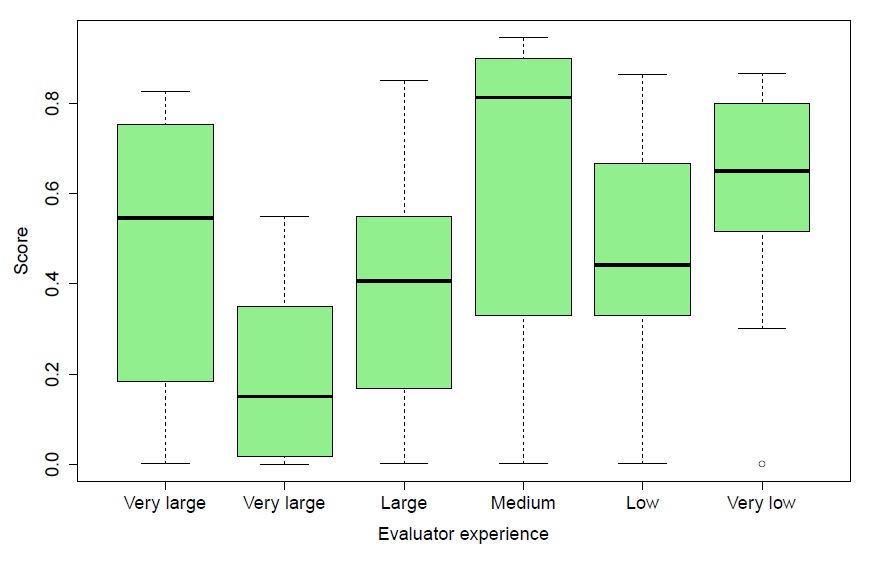}
\vspace{-0.4cm}
\caption{Subjective score - Medians of all evaluators. Boxplots of subjective scores for evaluate the ontology }
\label{}
\end{figure}

Concerning the medians of the objective scores, it was observed that the grades of the more experienced evaluators and less experienced ones were closer, i.e., the more grades that are used in the empirical study, the more the averages tend to be closer (see figures 6 and 7).

\begin{figure}[!htb]
\centering
\includegraphics[scale=0.5]{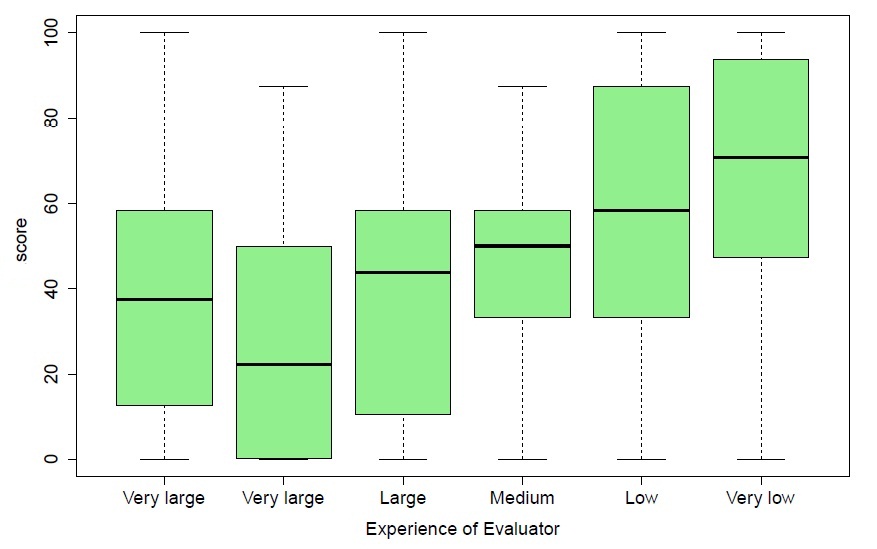}
\vspace{-0.4cm}
\caption{Objective score - Medians of all evaluators. Boxplots of scores for evaluate the ontologies}
\label{}
\end{figure}

Thus, the answer of research question 1.1 is: \textbf{Yes, when the methodology explains how to verify the question, approximates the grades of people with vast and little experience} (see Fig. 6 and Fig. 7).

Afterwards, the median of all the scores (subjective and objective separately) of all the evaluators were calculated to compare if the medians were the same. The next figure shows this comparison.

\begin{figure}[!htb]
\centering
\includegraphics[scale=0.47]{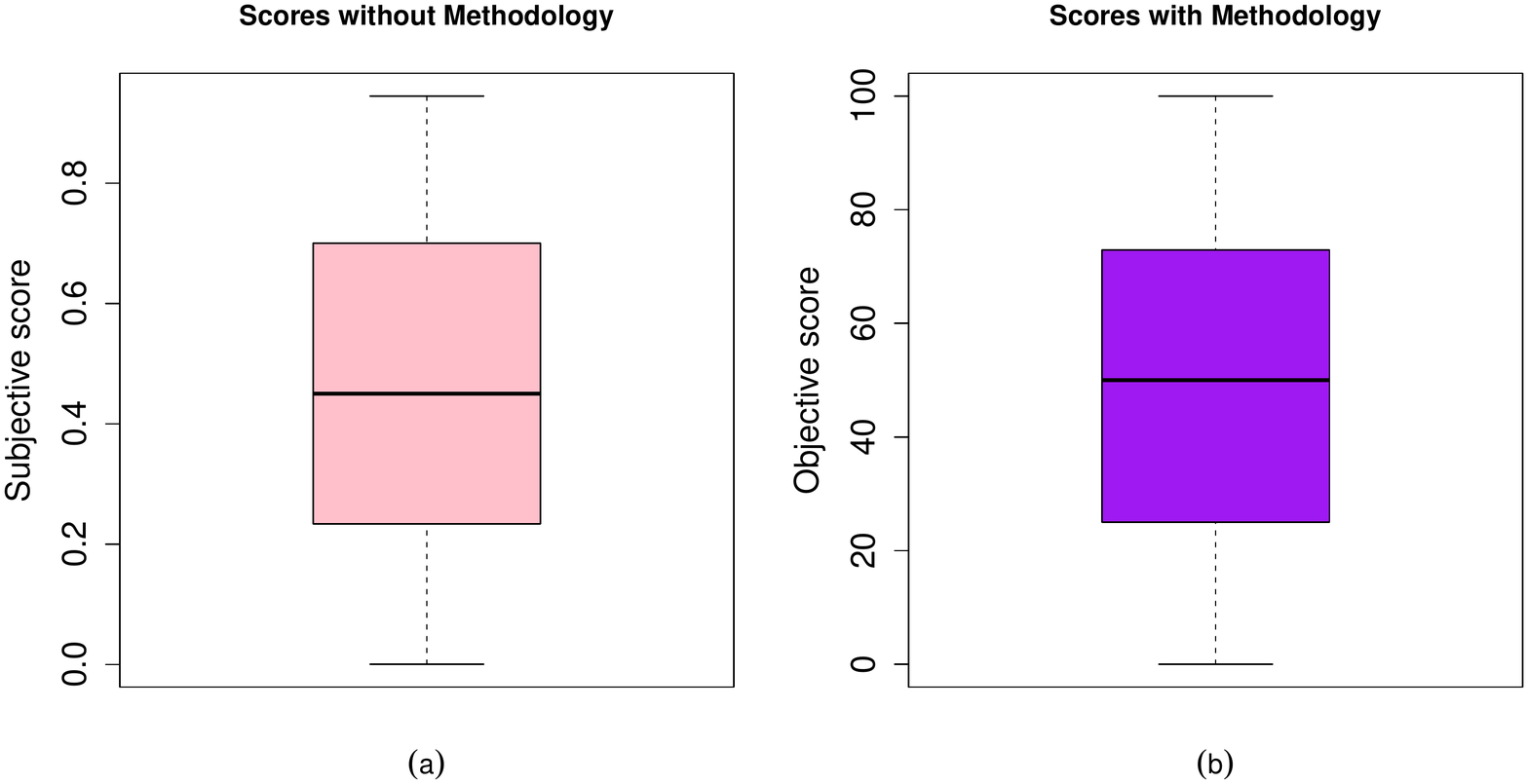}
\vspace{-1.3cm}
\caption{Medians of all (a) Subjective (b) and Objective scores}
\label{}
\end{figure}

When the evaluator does not use the methodology to evaluate the ontologies, he/she spends more time understanding the questions and how to verify them, i.e., more effort is required, because does not exist a step-by-step to guide him/her. Thus, the evaluator needs to understand both the criterion and the question. When the evaluator uses the FOCA Methodology, he/she answers the questions faster, there are steps to guide them and how to verify each question, i.e., less effort is required. The medians of the subjective and objective scores tend to be closer, i.e., the less effort required, the grades are the same. Thus, the answer of research question 1.2 is: \textbf{Yes, the evaluator requires less effort when he/she uses the methodology} (see Fig. 8).

To answer the research questions 2.1 and 2.2, a Linear Regression Model was made to observe if the subjective scores can be explained by objective scores provided by the FOCA Methodology. The variables involved are: the subjective scores (0.001,0.990), the evaluator's level of experience (very large, large, medium, low, very low) and the objective scores (0,100) of the GOALs (Substitute, Ontological Commitments, Intelligent Reasoning, Efficient Computation and Human Expression) considering the ontologies used in the empirical study (A, B, C, D).

First, the most appropriate distribution for the regression model response was investigated, as the subjective score variable assumes values in the interval (0, 1), the beta regression model proposed by \cite{r15} is an appropriate model. The beta distribution is commonly used to model random variates which assume values in the unit interval, such as rates, percentages and proportions. The beta density can show quite different shapes depending on the values of the parameters that index the distribution. Thus, we investigated several beta regression models, by using different statistical approaches to identify the best model, including models with varying dispersion (\cite{r17}). Finally, the systematic component of the model is defined by the following mathematical expression: 

\vspace{0.5cm}
\begin{mdframed}
\begin{footnotesize}
\begin{eqnarray*}
\hspace{-0.8cm}\frac{\exp \{\beta_1+ \beta_2{(Cov_S \times Sb)}_i + \beta_3 {(Cov_C \times Co)}_i +\beta_4{(Cov_R \times Re)}_i + \beta_5{(Cov_{Cp} \times Cp)}_i - \beta_6 LExp_i - \beta_7{Nl}_i\}}
{1+ {\exp \{\beta_1+ \beta_2{(Cov_S \times Sb)}_i + \beta_3 {(Cov_C \times Co)}_i + \beta_4{(Cov_R \times Re)}_i + \beta_5{(Cov_{Cp} \times Cp)}_i - \beta_6 LExp_i - \beta_7{ Nl}_i\}}}
\end{eqnarray*}
\end{footnotesize}
\end{mdframed}
\vspace{0.5cm}

It should be mentioned that the components of the regression model are the response variable  $y$; in this case, the subjective global scores for the ontology quality, the explanatory variables; the objective scores for each goal of the ontologies considered, the distribution of the probability of the response variables, the systematic mathematical expression and the coefficients of the explanatory variables. Considering the systematic expression above, it can be seen that the coefficients are $\beta_1, \beta_2,\beta_3,\beta_4,\beta_5,\beta_6$ and $\beta_7$. The response and explicative variable values are known, but the coefficient values are not known. Thus, these coefficients should be estimated. This is typically done using the maximum likelihood method. Thus, the estimated model is obtained using the observed subjective scores (or true values of response $y$) and using observed values of the objective scores (explicative variables). Table 18 presents the coefficients (first column) of its explanatory variables (second column), the estimated coefficients values (third column) and the p-Values that measure the coefficients significance, in other words, if the associated explanatory variables it is important to explain the score of ontology quality.

%\vspace{0.5cm}
%\begin{mdframed}
%\begin{eqnarray*}
%\hspace{-0.8cm} \widehat y_i=&\\
%&
%\hspace{-0.8cm}\frac{\exp \{-0.44+ 0.03{(Cov_S \times Sb)}_i + 0.02 %{(Cov_C \times Co)}_i + 0.01{(Cov_R \times Re)}_i + 0.02{(Cov_{Cp} %\times Cp)}_i - 0.66 LExp_i - 2.5{Nl}_i\}}
%{1+ {\exp \{-0.44+ 0.03{(Cov_S \times Sb)}_i + 0.02 {(Cov_C \times %Co)}_i + 0.01{(Cov_R \times Re)}_i + 0.02{(Cov_{Cp} \times Cp)}_i - %0.66 LExp_i - 2.5{ Nl}_i\}}}
%&
%\end{eqnarray*}
%\end{mdframed}
%\vspace{0.5cm}

\begin{table*}[ht]
	
	\caption{Estimated coefficients and the associated p-Values.}
	\begin{tabular}    {p{0.10\linewidth}p{0.20\linewidth}p{0.2\linewidth}p{0.2\linewidth}}
		\hline
		\textbf{Coefficients} & \textbf{Description} & \textbf{Estimative} & \textbf{p-Value}\\
		\hline
		$B_1$ & {Const} & $-0.44$  & 0.0000
		\\
		
		$B_2$ & {Cov $\times$ Sb} & 0.03 & 0.0001
		\\
		
		$B_3$ & {Cov $\times$ Co} & 0.02 & 0.0000
		\\
		
		$B_4$ & {Cov $\times$ Re} & 0.01 & 0.0000
		\\ 
		
		$B_5$ & {Cov $\times$ Cp} & 0.02 & 0.0013
		\\
	
		$B_6$ & {LExp} & $-0.66$ & 0.0549
		\\
		
		$B_7$ & Cov $\times$ Nl & $-25$ & 0.0005
		\\
		\hline
	\end{tabular}	

\end{table*}

As the mathematical expression has been estimated, we no longer need the observed score of the ontology quality (the subjective scores). We should only follow the step-by-step FOCA methodology and then a global or partial score of quality will be provided by the statistical model. The coefficient values of the estimated statistical model are very important. For example, $- 0.66 LExp_i$ implies that the more experienced the evaluator is, the stricter his/her evaluation of the ontology  quality will be, as the least will be the value of $\widehat y_i$. If the signal of a coefficient is negative, the bigger the explicative variable and the smaller the  $\widehat y_i$.  The effect of the $NI$ variables is more noteworthy ( it is 1 only if in some Goal it was impossible for the evaluator to answer all the questions). It can be observed that in the statistical model it appears as $- 2.5{ Nl}_i$. This means that if the ontology has some inconsistency and  the evaluator does not give an objective score to any of the goals of this ontology, the quality of the ontology will drop considerably. The magnitude of this decline is associated to the value of the coefficient estimates (in this case  $2.5$), and therefore it is higher when compared with the values of other coefficient estimates.

Finally, the adequacy of the fit of a regression model should be checked, which can be done by a diagnostic analysis, which in this case is the residual analysis. The residuals of a regression model is the difference between the estimated values of the response $\widehat y_i$ and the true values obtained from empirical study, $y_i$. If the residuals (the differences) are randomly scattered around zero, it is good evidence that the regression model is appropriate to describe the empirical study, and then the methodology can be validated. Figure 11 shows the plot of residuals compared to the observation indices, based on the proposal by \cite{r16} .

\begin{figure}[!htb]
\centering
\includegraphics[scale=0.4]{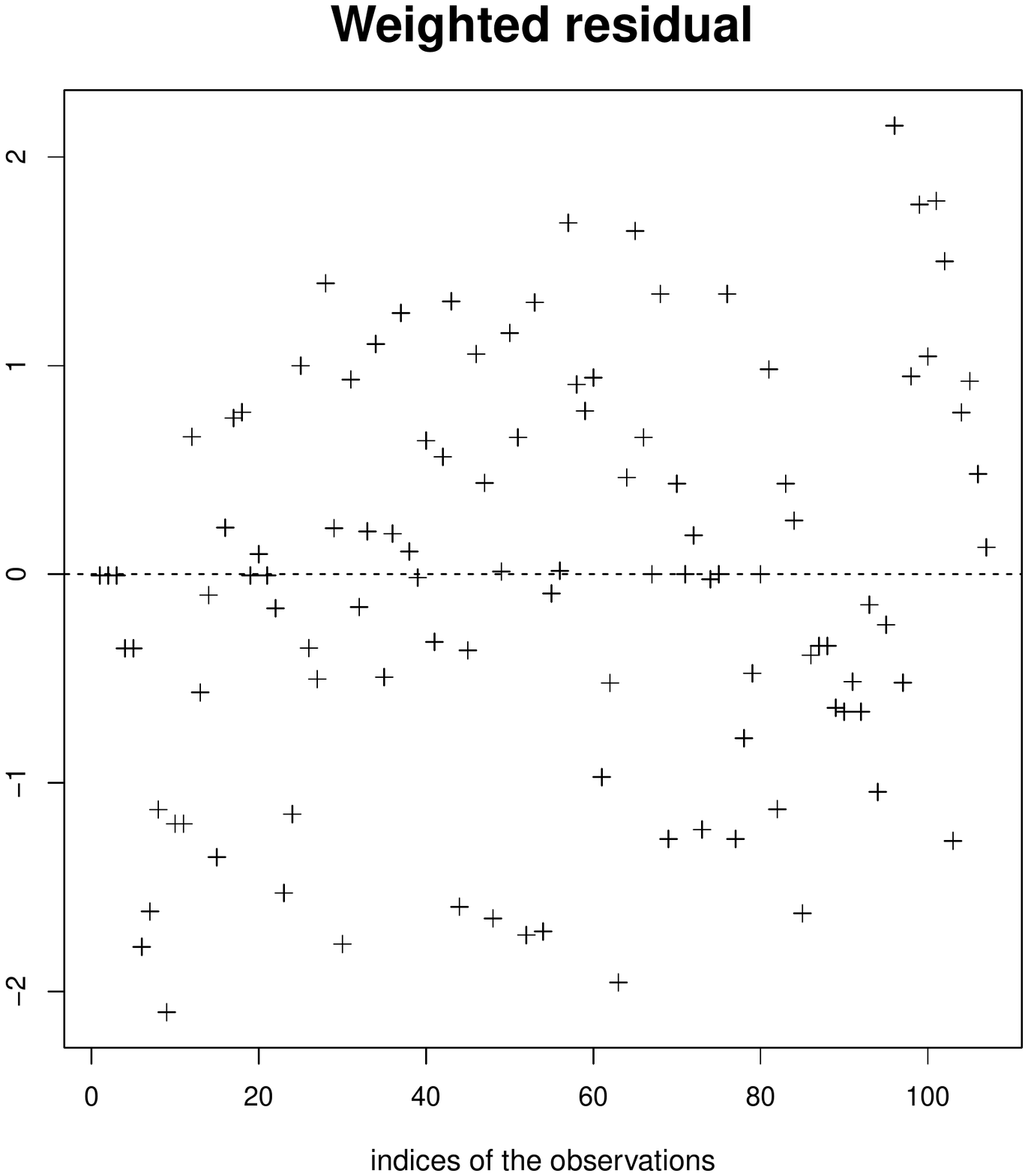}
\vspace{-0.4cm}
\caption{The residual plot}
\label{}
\end{figure}

Figure 9 shows that the difference (residual) is randomly scattered around zero. Thus, the answer for research questions 2.1 and 2.2 are: \textbf{Yes, it is possible to create an equation using the linear regression model to calculate the partial and total quality of an ontology}.

\subsection{Threats of validity}

The empirical study showed that the FOCA methodology is useful for all evaluators with different experiences and it produced an equation from the evaluator's grades, but there are some threats that can reduce the validity of the empirical study:

\begin{itemize}
	\item The empirical study used only one scenario to model the ontology, which could affect the grades given by the evaluators and hence the equation;
	\item The empirical study used only 4 ontologies to verify the quality, i.e., a small number of samples, which could also affect the grades and hence the equation;
	\item The fact that the evaluators were exhausted because they evaluated 4 ontologies without the method- ology and afterwards evaluated the 4 ontologies again, which may have affected the grades given by the evaluators and hence the equation;
	\item The number of participants per experience, because the empirical study had 2 samples per vast experience and 1 sample for the other experiences.  
\end{itemize}

\subsection{Discussion}

The participants of the empirical study reported that it was easier to evaluate the ontology with the step-by- step process than without it. More experienced people were able to evaluate it without using the methodology, but had to think what to do in each question, and took longer than with the FOCA Methodology. The less experienced people did not verify many questions when they did not use the methodology. However, when they used the FOCA Methodology, they verified all the questions.

The objective of the empirical study was achieved because the grades given by the more experienced eval- uators were close to the grades given by the less experienced evaluators. In addition, all the grades were used to create an equation where any evaluator may have a number between 0 and 1 that represents the quality of the ontology.

There are some observations to be made concerning this Methodology. It should be mentioned that this study does not include all the questions presented in \cite{r9} because it would be an exhaustive process to use them all. Thus, only the main questions were chosen. Some questions were devised by the author and others were adapted. Note that there are no variables for the Human Expression role in the mathematical expression of the quality score, nonetheless it is important to explain that this score is implicit in the equation, because the human expression refers to the evaluator's expertise and the ability to answer all the questions.

\section{Conclusions}

This work presented a new methodology for ontology evaluation, called FOCA. The methodology consists of three main steps: Ontology Type Verification, Questions Verification and Quality Verification. Most studies about methodologies for ontology evaluation only present the criteria, and do not provide guidelines for the evaluator concerning their approaches. This causes the evaluation to be inaccurate, and also depends on the evaluators' experience. Thus, there was a need to create this methodology.

Users of ontology communities now have a step-by-step guide to evaluate ontologies. Moreover, the evaluators know how the ontology fullfils each role of the knowledge representation, with less effort, regardless of his/her experience and also has a numerically precise diagnosis. However, this methodology has some limitations. For instance, the lack of questions which makes the ontology evaluation more distinct according to the types of task or domain and application ontology (only two questions make this difference). Another limitation is that, evaluating two or more ontologies, checking the type, answering thirteen questions and calculating the quality makes the process rather exhausting. Finally, the questions are still slightly subjective (such as Q6, Q7, Q8 and Q9) which can affect the final score.

In future research, aiming to eliminate the limitations and the possibility of inserting Top Ontology will be studied, including more questions about this type. Moreover, the possibility of a semi-automated or automated methodology will be investigated so as not to exhaust the evaluator. It is hoped that these procedures will improve the construction of the statistical model that supports the FOCA methodology.

\end{document}